\documentclass[conference]{IEEEtran}

\makeatletter
\def\ps@IEEEtitlepagestyle{%
  \def\@oddfoot{\mycopyrightnotice}%
  \def\@evenfoot{}%
}
\def\mycopyrightnotice{%
  {\footnotesize XXX-X-XXXX-XXXX-X/XX/\$XX.00~\copyright~20XX IEEE\hfill}% <--- Change here
  \gdef\mycopyrightnotice{}
}

\usepackage{blindtext}
\usepackage{eso-pic}
\IEEEoverridecommandlockouts
\usepackage{cite}
\usepackage{amsmath,amssymb,amsfonts}
\usepackage{algorithm}
\usepackage{algpseudocode}
\usepackage{graphicx}
\usepackage{textcomp}
\usepackage{xcolor}
\usepackage{hyperref}
\def\BibTeX{{\rm B\kern-.05em{\sc i\kern-.025em b}\kern-.08em
    T\kern-.1667em\lower.7ex\hbox{E}\kern-.125emX}}
    
\usepackage{eso-pic}
\newcommand\AtPageUpperMyright[1]{\AtPageUpperLeft{%
 \put(\LenToUnit{0.17\paperwidth},\LenToUnit{-2cm}){%
     \parbox{0.9\textwidth}{\raggedleft\fontsize{8}{11}\selectfont #1}}%
 }}%
\newcommand{\conf}[1]{%
\AddToShipoutPictureBG*{%
\AtPageUpperMyright{#1}
}
}

\begin{document}
\title{\vspace*{1cm} Privacy-Preserving Automated Rosacea Detection Based on Medically Inspired
Region of Interest Selection\\
}

\author{\IEEEauthorblockN{Chengyu Yang}
\IEEEauthorblockA{\textit{Department of Computer Science} \\
\textit{New Jersey Institute of Technology}\\
Newark, USA \\
cy322@njit.edu}
\and
\IEEEauthorblockN{Rishik Reddy Yesgari}
\IEEEauthorblockA{\textit{Department of Computer Science} \\
\textit{New Jersey Institute of Technology}\\
Newark, USA \\
ry248@njit.edu}
\and
\IEEEauthorblockN{Chengjun Liu}
\IEEEauthorblockA{\textit{Department of Computer Science} \\
\textit{New Jersey Institute of Technology}\\
Newark, USA \\
cliu@njit.edu}
}

\maketitle
\conf{\textit{  Proc. of International Conference on Electrical and Computer Engineering Researches (ICECER 2025) \\ 
6-8 December 2025, Antananarivo - Madagascar}}
\begin{abstract}
Rosacea is a common but underdiagnosed inflammatory skin condition that primarily affects the central face and presents with subtle redness, pustules, and visible blood vessels. Automated detection remains challenging due to the diffuse nature of symptoms, the scarcity of labeled datasets, and privacy concerns associated with using identifiable facial images. A novel privacy-preserving automated rosacea detection method inspired by clinical priors and trained entirely on synthetic data is presented in this paper. Specifically, the proposed method, which leverages the observation that rosacea manifests predominantly through central facial erythema, first constructs a fixed redness-informed mask by selecting regions with consistently high red channel intensity across facial images. The mask thus is able to focus on diagnostically relevant areas such as the cheeks, nose, and forehead and exclude identity-revealing features. Second, the ResNet-18 deep learning method, which is trained on the masked synthetic images, achieves superior performance over the full-face baselines with notable gains in terms of accuracy, recall and F1 score when evaluated using the real-world test data. The experimental results demonstrate that the synthetic data and clinical priors can jointly enable accurate and ethical dermatological AI systems, especially for privacy sensitive applications in telemedicine and large-scale screening. The code and data are available at: \url{https://osf.io/d4nz9/view_only=487c1130a1bb4613bb796aea5adfe116}
\end{abstract}

\begin{IEEEkeywords}
Rosacea detection, Deep Learning, Masked Images, Medical Prior, Medical Imaging.
\end{IEEEkeywords}

\section{Introduction}
Rosacea is a chronic inflammatory skin condition that primarily affects the central face, including the cheeks, nose, and forehead \cite{b4}. Clinically, it manifests through persistent erythema (redness), papules, pustules, and visible blood vessels. While rosacea is not life-threatening, its visible nature can lead to significant psychosocial distress and a reduced quality of life. Despite its prevalence, automated detection of rosacea remains challenging due to subtle and diffuse visual symptoms, a lack of annotated datasets, and growing concerns around patient privacy when using full facial images.

The first challenge stems from the nature of the disease itself: the redness patterns associated with rosacea are often mild and overlapping with other skin conditions or natural variations in skin tone. As a result, conventional computer vision methods may struggle to capture reliable diagnostic features. Second, and more critically in practice, using full facial images raises serious privacy issues. Facial imagery contains strongly identifying features such as the eyes, mouth – elements that are not necessary for rosacea diagnosis but pose ethical and regulatory barriers to dataset collection and model deployment. Finally, publicly available datasets containing labeled rosacea images are extremely limited. Data scarcity makes training deep neural networks prone to overfitting and hampers their ability to generalize to diverse populations.

In this work, we address these challenges through a novel, privacy-preserving rosacea detection framework that leverages synthetic inspired priors. Our method is based on the clinical observation that rosacea predominantly manifests as redness in the central facial area together with the forehead. We construct a fixed redness-informed mask by averaging pixelwise red channel intensities across a set of facial images and selecting regions with consistently high red values. This mask emphasizes the cheeks, forehead and nasal bridge – areas commonly affected by rosacea – while excluding identity – revealing features such as the eyes and mouth. The same mask is applied uniformly to all input images, ensuring consistency and privacy preservation across the dataset.

Due to the scarcity of real annotated images, we use entirely
synthetic facial data for both training and validation. These synthetic images include rosacea-positive and rosacea-negative (healthy) examples, generated by the GAN to simulate realistic skin textures and redness distributions. This approach allows us to create a balanced dataset that captures clinically relevant variations in appearance, while eliminating the need to collect sensitive patient data.

A ResNet-18 \cite{b7} classifier is then trained on the masked synthetic images to perform binary classification between rosacea and normal skin. Our results show that this masked, redness-focused approach significantly outperforms baseline models trained on unmasked full-face images, with notable improvements in accuracy, recall and F1 score when test on real data. These findings highlight the dual benefit of our framework, enhancing diagnostic performance while preserving patient privacy.

In summary, our contributions are three-fold. First, we introduce a clinically informed, fixed facial mask that highlights diagnostically relevant regions while obscuring identity, enabling privacy-aware dermatological analysis. Second, we propose a novel rosacea detection framework trained entirely on synthetic data, eliminating reliance on real facial images which are hard to collect. Third, we demonstrate that our method achieves superior classification performance compared to full-face baselines when tested on real-world patients’ facial images, validating the effectiveness of incorporating medical priors into privacy-preserving vision models. 

Our results suggest that ethical and accurate AI systems for dermatology are not mutually exclusive. By combining generated disease images with clinical insights and privacy-preserving design, we offer a promising direction for automated skin disease detection – especially in contexts such as telemedicine, mobile health, and large-scale screening, where privacy and scalability are paramount.

\section{Related Work}

\subsection{Automated Skin Disease Detection}

Deep learning has advanced the field of dermatological image analysis, with convolutional neural networks (CNNs) achieving dermatologist-level performance on various classification tasks. Esteva et al. \cite{b3} demonstrated the potential of CNNs in skin cancer classification, catalyzing broader adoption of deep learning in medical imaging. Tschandl et al. \cite{b11} further contributed with the HAM10000 dataset, enabling standardized evaluation across multiple skin conditions. However, most existing work focuses on dermoscopic or lesion-centric images rather than full-face conditions like rosacea, which presents diffuse, subtle patterns over the central face.

\subsection{Privacy in Medical Computer Vision}

Preserving patient privacy in facial analysis has received growing attention, particularly in the context of healthcare and biometric data. Approaches such as facial obfuscation, blurring, and region masking have been explored to remove identity features while retaining task-relevant information \cite{b15}. In dermatology, identity obfuscation often conflicts with the need to retain diagnostic information. Our approach circumvents this by designing a redness-informed mask based on clinical priors, targeting diagnostic areas while eliminating identity-revealing features.

\subsection{Synthetic Data for Medical Imaging}

Synthetic data generation has become increasingly important for addressing data scarcity in medical domains. \cite{b1} used generative adversarial networks (GANs) to augment liver lesion datasets and demonstrated improved classification accuracy. Due to the scarcity of real patients’ rosacea images \cite{b13}, we leverage the synthetic rosacea images with GAN models \cite{b10}. This generated dataset has been used to enhance performance and generalization ability for some automatic rosacea detection methods \cite{b14}.

\subsection{Clinically Informed AI}

Integrating clinical domain knowledge—such as known anatomical patterns or symptom localization—into model design helps guide the deep neural network’s focus toward diagnostically relevant features, thereby improving both interpretability and classification performance. For example, attention-based models have been used to highlight disease-relevant regions in chest X-rays \cite{b6} and retinal images \cite{b12}. In dermatology, clinically driven priors like redness, texture, or lesion location remain underutilized. We address this gap by incorporating a redness-informed spatial mask, derived from statistical red-channel intensity distributions, to guide the model to focus on the regions of the face that are most relevant to this disease without requiring pixel-level annotations.

\section{PRIVACY-PRESERVING AUTOMATED ROSACEA DETECTION BASED ON MEDICALLY INSPIRED REGION OF INTEREST SELECTION}

In this section, we present a novel privacy-preserving automated rosacea detection using a privacy enhancing region of interest selection approach and a deep learning model. First, we identify region of interest (ROI) based on established medical knowledge of rosacea, which typically shows symptoms of redness around areas such as the cheeks, nose, and forehead. The ROI is determined by analyzing red channel intensity distributions across facial images. Second, we train a deep learning model using facial images where only the selected ROI is retained, while the remaining pixels are masked out. This approach preserves diagnostically relevant features while obscuring identity-related information, supporting both effective classification and patient privacy.

\subsection{Medically Inspired Region of Interest Selection for Enhancing Patient Privacy}
We now present our method for deriving the region of interest (ROI) for rosacea detection, which is inspired by the established clinical knowledge and it enhances patient privacy by obscuring identity-revealing features. Specifically, one of the most salient and consistent visual symptoms of rosacea is facial redness (erythema), typically concentrated around the cheeks, nose, and forehead \cite{b2}. To capture these diagnostically relevant areas, we aim to identify facial regions with the highest likelihood of exhibiting redness.

Rather than relying on manual annotation to select these regions, we utilize the red component image from the RGB facial images, which directly reflects the intensity values associated with erythema. To systematically extract the disease-relevant regions, we first compute the mean face image using all the rosacea-positive samples from the training set. We then extract the red component image from this mean image and select the top $t$ percent of the pixels with the highest red intensity values. The parameter $t$ serves as a threshold that controls the size of the selected region and can be either manually specified or empirically determined through the validation experiments. 

This thresholding approach yields a fixed redness-informed mask that predominantly emphasizes the cheeks, nasal bridge, and forehead areas clinically recognized as common sites of rosacea manifestation. Furthermore, we have noticed that the regions typically associated with facial identity, such as the eyes and mouth, are naturally excluded by our masking process. This is because these areas tend to have low red channel intensities and thus fall below the threshold used for selecting the top $t$ percent of the red pixels. As a result, the final redness-informed mask inherently omits identity-revealing features while focusing on diagnostically relevant regions, offering both clinical utility and privacy preservation without requiring explicit manual exclusion.

The final mask thus serves a dual purpose: it highlights diagnostically informative regions aligned with clinical observations while simultaneously obfuscating identity-revealing features. This enables effective model training on privacy-preserving inputs without sacrificing diagnostic performance. Let the training set have $n$ rosacea positive samples and each image in the training set is an RGB image with height $h$ and width $w$. The total number of values per image is $3hw$, accounting for the three-color channels. In the red channel alone, there will be $hw$ values. The procedures are summarized in Algorithm \ref{alg:roi}.

\begin{algorithm}
\caption{Region of Interest (ROI) Selection}\label{alg:roi}
\begin{algorithmic}[1]
\Require 
\\$t\in (0,100]$ \Comment{cutoff at top $t$\%}
\\ $X\in \mathbb{R}^{3hw \times n} $ \Comment{rosacea positive samples from the training set} 
\Ensure ROI (Region of Interest) \Comment{a binary mask}
\State $m \gets \bar{X}$\Comment{Get the mean face from rosacea positive samples }
\State $r \gets m[0,\ :\ ,\ :\ ]$ \Comment{Extract the red channel from the RGB mean face}
\State $\dot r \gets sort(r)$\Comment{Sort values in $r$ in decreasing order}
\State $s \gets \dot r[\lceil t\% \rceil \cdot w \cdot h]$\Comment{Get the threshold value $s$}

 \For{each pixel $i$ in $r$}
    \If{$i > s$}
        \State $i \gets 1$
    \Else
        \State $i \gets 0$
    \EndIf
\EndFor
\State \Return $r$
\end{algorithmic}
\end{algorithm}

\begin{figure}
    \centering
    \includegraphics[width=0.99\linewidth]{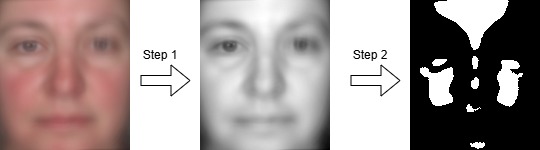}
    \caption{The process of selecting the region of interest for rosacea detection and enhancing patient privacy. The first image is the mean RGB image of the rosacea positive samples from the training set. The image in the middle is the red component image from the mean RGB image and Step 1 applies it to derive a threshold corresponding to the top t\% of the pixels. The last one shows the region of interest defined in Step 2 by comparing the pixel intensity values in the red componet image with the threshold: pixels with intensity values above the threshold are set to 1, otherwise 0. }
    \label{example}
\end{figure}

Fig.\ref{example} shows the process of selecting the region of interest for rosacea detection and enhancing patient privacy. The first image represents the mean RGB face image computed from all the rosacea-positive samples in the training set. In Step 1, the red component image is extracted from this mean image to isolate the redness-related features. The image in the middle in Fig.\ref{example} is the red component image. A threshold is then defined based on the top $t$\% of the red intensity values. Step 2 finally defines the region of interest by comparing the pixel intensity values in the red component image with the threshold: pixels with intensity values above the threshold are set to 1, and all others are set to 0. As a result, this binary mask, which is the last one shown in Fig.\ref{example}, highlights the regions most indicative of rosacea-related erythema.

\subsection{Privacy-Preserving Automated Rosacea Detection Using a Deep Learning Model}

\begin{figure}
    \centering
    \includegraphics[width=0.99\linewidth]{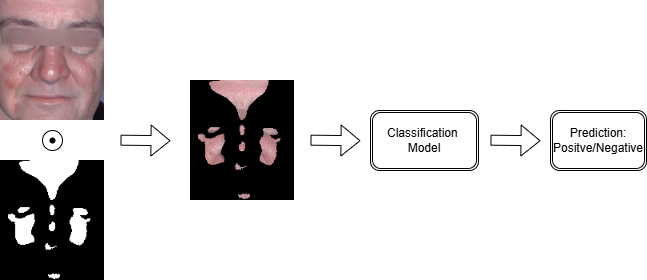}
    \caption{The system architecture of the proposed privacy-preserving automated rosacea detection using the privacy enhancing region of interest selection approach and a deep learning model. The first column shows an example image and the ROI mask. The second column displays the masked image that not only captures the most discriminating medical features for rosacea detection but also enhances patient privacy by obscuring identifiable information. The remaining two columns are the deep learning model for rosacea detection.}
    \label{flow}
\end{figure}

We now present our method for privacy-preserving automated rosacea detection using the privacy enhancing region of interest selection approach and a deep learning model. In particular, after selecting the region of interest, a deep learning model is trained using images masked using a mask as shown in Fig.\ref{example}. Let an image from the dataset be $x$ and the mask be $m$. The process of applying the mask to the image is simply element-wise matrix multiplication:

\begin{equation}
 x = x \odot m
\end{equation}

where $\odot$ represents element-wise matrix multiplication.

Fig.\ref{flow} shows the system architecture of the proposed privacy- preserving automated rosacea detection using the privacy enhancing region of interest selection approach and a deep learning model. Specifically, the first column in Fig.\ref{flow} shows an example image and the ROI mask. The second column in Fig.\ref{flow} displays the masked image that not only captures the most discriminating medical features for rosacea detection but also enhances patient privacy by obscuring identifiable information. The remaining two columns are the deep learning model for rosacea detection.

For rosacea detection, we apply the ResNet-18 \cite{b7} deep learning method, which is a lightweight yet effective convolutional neural network that incorporates residual learning to address the vanishing gradient problem commonly encountered in deeper models. With 18 layers, the ResNet-18 deep learning method strikes a balance between the representational capacity and the computational efficiency, which makes it suitable for medical imaging tasks where the datasets are often limited in size. In the context of rosacea detection, combining the ResNet-18 deep learning method with the masked region of interest enhances the model’s focus on diagnostically relevant facial areas while reducing the background noise and irrelevant features. This targeted input allows the network to concentrate its learning capacity on subtle but meaningful variations in skin texture and redness patterns associated with rosacea. The residual architecture further ensures stable and efficient training, thus enabling the model to learn complex patterns from the masked inputs more effectively.

\begin{figure*}
    \centering
    \includegraphics[width=\textwidth]{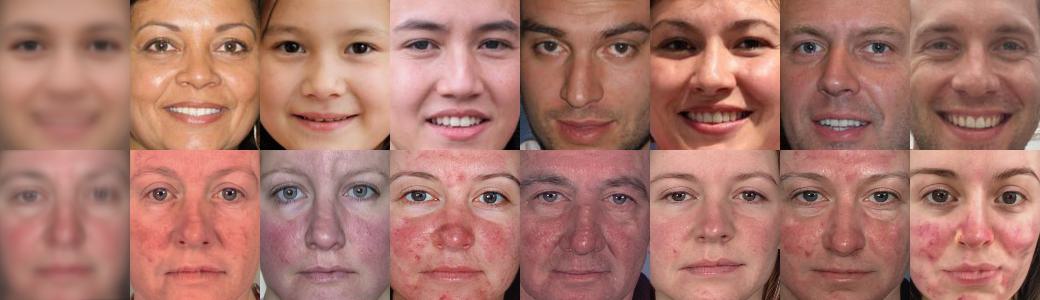}
    \caption{The first column shows the mean images of the two training datasets from the rosacea negative and positive classes, respectively. The remaining seven columns display seven pair of random example images from the two training datasets corresponding to the normal people and those with rosacea, respectively.
}
    \label{fig2}
\end{figure*}

\section{Experiments}

In this section, we describe the experiments conducted to evaluate the effectiveness of our proposed method. We first present the dataset used in our study and detail the preprocessing steps applied. Next, we outline the experimental setup and report the results. The findings demonstrate that our approach achieves superior performance when compared to baseline methods, highlighting its potential for accurate and privacy-preserving rosacea detection.

\subsection{Dataset}

We leverage the same dataset as used in \cite{b13}, \cite{b14} to simulate real-world scenarios where training data is limited and difficult to collect, while also safeguarding patient privacy. To further enhance data quality, we align the facial images and remove the background by cropping and centering the faces. This preprocessing step helps eliminate irrelevant background information and facilitates the application of the mask.

\subsubsection{Training and Validation}
We utilize generated images for both rosacea-positive and rosacea-negative cases. As illustrated in Figure 3, the frontal face images with rosacea are synthesized using a GAN framework \cite{b10} \cite{b5}. Specifically, we employ 300 generated rosacea-positive images, dividing them into a training set of 250 images and a validation set of 50 images. For rosacea-negative cases, we use 600 frontal face images generated with StyleGAN \cite{b8}, of which 500 are allocated to the training set and 100 to the validation set.

\subsubsection{Testing}

To evaluate the performance of the proposed automatic rosacea detection method on real-world data, we use 50 real frontal face images with rosacea, sourced from platforms such as Kaggle, DermNet, and the National Rosacea Society. These images are selected based on minimum resolution criterion of 200×200 pixels. Each face is manually aligned, cropped and resized to match the dimensions of the training and validation images (150 × 130 × 3). Additionally, we include 150 real frontal face images without rosacea from the CelebA dataset [9] as negative samples in the test set.

\subsection{Determine the Threshold for the Region of Interest}

To determine the threshold for the selection of the regions of interest, we iterate the top percentage from 20 to 35. For each threshold selected, we apply the mask derived from this threshold to all the images in the dataset. Then a classification model was trained on the training set. We select the threshold that gives the best overall performance for testing. We use 4 metrics for now to select the threshold that gives the best validation performance.

To be more specific, we use accuracy, recall rate, precision rate, and F1 score. For a single method, all these four metrics are the higher, the better. The definitions of these four metrics are given below, where TP stands for the number of true positive predictions, TN stands for the number of true negative predictions, FP stands for the number of false positive predictions and FN stands for the number of false negative predictions.

\begin{equation}
\text{Accuracy} = \frac{\text{TP} + \text{TN}}{\text{TP} + \text{TN} + \text{FP} + \text{FN}}
\end{equation}

\begin{equation}
\text{Recall} = \frac{\text{TP}}{\text{TP} + \text{FN}}
\end{equation}

\begin{equation}
\text{Precision} = \frac{\text{TP}}{\text{TP} + \text{FP}}
\end{equation}

\begin{equation}
\text{F1-Score} = 2 \cdot \frac{\text{Precision} \cdot \text{Recall}}{\text{Precision} + \text{Recall}}
\end{equation}

During our experiments, we observed that as the number of training epochs increases, the performance gap between different threshold values gradually diminishes. This phenomenon is likely due to overfitting, as both the training and
validation sets consist entirely of synthetic data. To better capture the sensitivity of the model to different threshold values, we therefore select the optimal threshold after training the classification model for only three epochs, where the effects of the threshold are more pronounced. For final testing, however, we train the model for the full 30 epochs to ensure optimal performance.

\begin{figure}
    \centering
    \includegraphics[width=0.99\linewidth]{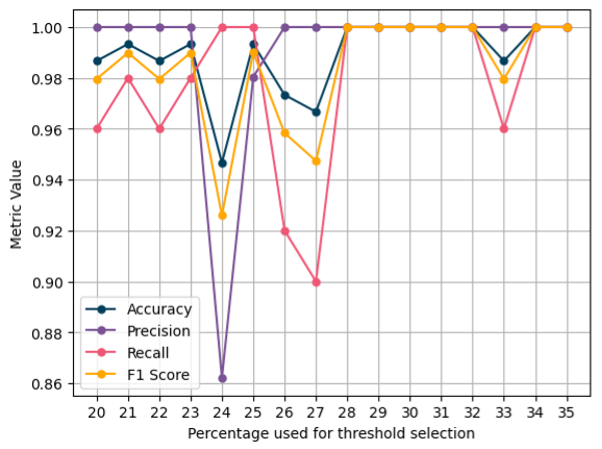}
    \caption{Validation performance for different shresholds}
    \label{validation}
\end{figure}

\begin{figure}
    \centering
    \includegraphics[width=0.8\linewidth]{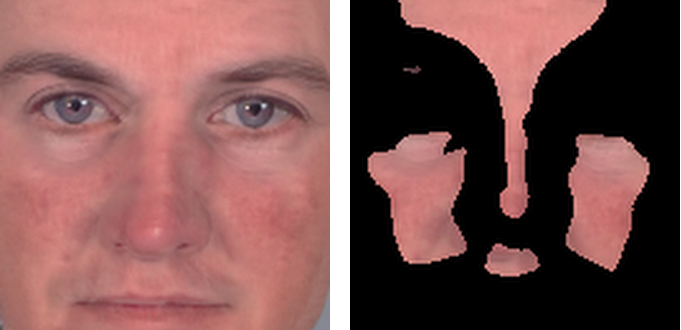}
    \caption{Identity Protection Effect After Applying the Mask}
    \label{privacy}
\end{figure}

As shown in Fig.\ref{validation}, the model achieves peak performance on the validation set when the threshold corresponds to the top 28\% of red channel values. To balance performance with privacy preservation, we select a slightly higher threshold of 29\%. This choice maintains the best validation performance, while keeping the unmasked facial region small, thereby enhancing identity protection.

\subsection{Testing Performance on Unseen Data}

As shown in Table~\ref{tab1}, our method significantly outperforms the baseline approach that uses unmasked images. Although using original images achieves 100\% precision, it suffers from a low recall of just 0.34, indicating a substantial failure to identify true positive cases. Specifically, as reported in Table~\ref{tab3}, it correctly detects only 17 out of 50 rosacea patients. This low recall suggests that many cases may go undiagnosed, missing the opportunity for early intervention. In contrast, our proposed method using the masked images not only achieves nearly perfect precision (97.78\%, with only 1 false alarm out of 150 negative samples), it misses only 6 out of 50 patients, significantly improving the detection rate (recall rate) compared to the method using unmasked images. This heightened sensitivity enhances the likelihood of timely diagnosis and treatment, which is crucial for favorable clinical outcomes in rosacea care. Moreover, as illustrated in Figure 5, our approach preserves patient identity, demonstrating that improved performance and privacy protection are not mutually exclusive.

\begin{table}[htbp]
\caption{The comparison of performance metrics on the classification model}
\begin{center}
\begin{tabular}{|l|c|c|c|c|}
\hline
\textbf{Images Used} & \textbf{Accuracy} & \textbf{Recall} & \textbf{Precision} & \textbf{F1} \\
\hline
Original Images & 0.8350 & 0.3400 & 1.0000 & 0.5075 \\
\hline
\textbf{Masked Images} & \textbf{0.9550} & \textbf{0.8200} & \textbf{1.0000} & \textbf{0.9011} \\
\hline
\end{tabular}
\label{tab1}
\end{center}
\end{table}

\begin{table}[htbp]
\caption{Test results: Confusion matrix of our method for Testing}
\begin{center}
\begin{tabular}{|l|c|c|}
\hline
 & \textbf{Predicted Normal} & \textbf{Predicted Rosacea} \\
\hline
\textbf{Actual Normal} & TN = 149 & FP = 1 \\
\hline
\textbf{Actual Rosacea} & FN = 6 & TP = 44 \\
\hline
\end{tabular}
\label{tab2}
\end{center}
\end{table}

\begin{table}[htbp]
\caption{Test results: Confusion matrix of using unmasked images}
\begin{center}
\begin{tabular}{|l|c|c|}
\hline
 & \textbf{Predicted Normal} & \textbf{Predicted Rosacea} \\
\hline
\textbf{Actual Normal} & TN = 150 & FP = 0 \\
\hline
\textbf{Actual Rosacea} & FN = 33 & TP = 17 \\
\hline
\end{tabular}
\label{tab3}
\end{center}
\end{table}

\section{Conclusion}

In this work, we presented a novel, privacy-preserving
method for automated rosacea detection inspired by clinical
priors and powered by synthetic data. By leveraging redness-
informed region-of-interest (ROI) masking based on statistical
analysis of red channel intensities, our method effectively focuses on diagnostically relevant facial regions such as the cheeks, nose, and forehead while obfuscating identity-revealing features. This enables high diagnostic performance without compromising patient privacy.

Extensive experiments on synthetic training data and real- world test images demonstrate that our approach significantly outperforms baseline models using unmasked images. Notably, our method improves recall from 0.34 to 0.82 while maintaining perfect precision, underscoring its potential for reducing missed diagnoses in clinical or telehealth settings. Moreover, the use of synthetic data eliminates the need for sensitive real patient images during training, addressing ethical and regulatory concerns.

Overall, our findings suggest that privacy and performance are not mutually exclusive in medical AI. By integrating domain knowledge, synthetic data generation, and targeted masking, we demonstrate a scalable, ethical, and effective approach to skin disease detection. Future work may explore dynamic or personalized masks, real-time deployment in telemedicine platforms, and extension to other dermatological conditions requiring visual analysis.

\vspace{12pt}


\begin{thebibliography}{00}
\bibitem{b1} C. Baur, S. Albarqouni, and N. Navab, “Generating highly realistic images of skin lesions with GANs,” Springer, 2018, pp. 260–267.
\bibitem{b2} H. Binol, A. Plotner, J. Sopkovich, B. Kaffenberger, M. K. K. Niazi, and M. N. Gurcan, “Ros-net: A deep convolutional neural network for automatic identification of rosacea lesions,” Skin Research and Technology, Vol. 26, No. 3, 2020, pp. 413–421.
\bibitem{b3} A. Esteva, B. Kuprel, R. A. Novoa, J. Ko, S. M. Swetter, H. M. Blau, and S. Thrun, “Dermatologist-level classification of skin cancer with deep neural networks,” Nature, Vol. 542, No. 7639, 2017, pp. 115–118.
\bibitem{b4} L. Ge, Y. Li, Y. Wu, Z. Fan, and Z. Song, “Differential diagnosis of rosacea using machine learning and dermoscopy,” Clinical, Cosmetic and Investigational Dermatology, 2022, pp. 1465–1473.
\bibitem{b5} I. Goodfellow, J. Pouget-Abadie, M. Mirza, B. Xu, D. Warde-Farley, S. Ozair, A. Courville, and Y. Bengio, “Generative adversarial networks,” Communications of the ACM, Vol. 63, No. 11, 2020, pp. 139–144.
\bibitem{b6} Q. Guan and Y. Huang, “Multi-label chest X-ray image classification via category-wise residual attention learning,” Pattern Recognition Letters, Vol. 130, 2020, pp. 259–266.
\bibitem{b7} K. He, X. Zhang, S. Ren, and J. Sun, “Deep residual learning for image recognition,” in Proc. IEEE Conf. Computer Vision and Pattern Recognition, 2016, pp. 770–778.
\bibitem{b8} T. Karras, S. Laine, and T. Aila, “A style-based generator architecture for generative adversarial networks,” in Proc. IEEE/CVF Conf. Computer Vision and Pattern Recognition, 2019, pp. 4401–4410.
\bibitem{b9} Z. Liu, P. Luo, X. Wang, and X. Tang, “Deep learning face attributes in the wild,” in Proc. IEEE Int. Conf. Computer Vision, 2015, pp. 3730–3738.
\bibitem{b10} A. Mohanty, A. Sutherland, M. Bezbradica, and H. Javidnia, “Towards synthetic generation of clinical rosacea images with GAN models,” in Proc. Irish Signals and Systems Conf., 2022, pp. 1–5.
\bibitem{b11} P. Tschandl, C. Rosendahl, and H. Kittler, “The HAM10000 dataset, a large collection of multi-source dermatoscopic images of common pigmented skin lesions,” Scientific Data, Vol. 5, No. 1, 2018, pp. 1–9.
\bibitem{b12} Z. Wang, Y. Yin, J. Shi, W. Fang, H. Li, and X. Wang, “Zoom-in-net: Deep mining lesions for diabetic retinopathy detection,” in Medical Image Computing and Computer Assisted Intervention – MICCAI 2017, Vol. 20, Springer, 2017, pp. 267–275.
\bibitem{b13} C. Yang and C. Liu, “Increasing rosacea awareness among population using deep learning and statistical approaches,” in Proc. Int. Conf. Medical Imaging and Computer-Aided Diagnosis, Springer, 2024, pp. 110–119.
\bibitem{b14} C. Yang and C. Liu, “Interpretable automatic rosacea detection with whitened cosine similarity,” arXiv preprint, arXiv:2504.08073, 2025.
\bibitem{b15} K. Yang, J. H. Yau, L. Fei-Fei, J. Deng, and O. Russakovsky, “A study of face obfuscation in ImageNet,” in Proc. Int. Conf. Machine Learning, PMLR, 2022, pp. 25313–25330.
\end{thebibliography}
\end{document}